\documentclass[runningheads]{llncs}
\usepackage[T1]{fontenc}
\usepackage{color, soul}
\usepackage{subfigure}
\usepackage{amsmath} 
\usepackage{booktabs}
\usepackage{multirow}
\usepackage{hyperref}

\usepackage{graphicx}
\usepackage{bbm}
\usepackage{amssymb} 
\begin{document}
%
\title{TE-SSL: Time and Event-aware Self Supervised Learning for Alzheimer's Disease Progression Analysis
}
%
%
\author{Jacob Thrasher\inst{1} \and
Alina Devkota\inst{1} \and
Ahmed P. Tafti\inst{2} \and
Binod Bhattarai\inst{3} \and
Prashnna Gyawali\inst{1},
for the Alzheimer's Disease Neuroimaging Initiative*
}

\authorrunning{Thrasher et al.}
%
\institute{West Virginia University, Morgantown, USA \and
University of Pittsburgh, Pittsburgh, USA \and
University of Aberdeen, Aberdeen, UK}
%
\maketitle              

\begin{abstract}
Alzheimer's Disease (AD) represents one of the most pressing challenges in the field of neurodegenerative disorders, with its progression analysis being crucial for understanding disease dynamics and developing targeted interventions. Recent advancements in deep learning and various representation learning strategies, including self-supervised learning (SSL), have shown significant promise in enhancing medical image analysis, providing innovative ways to extract meaningful patterns from complex data. Notably, the computer vision literature has demonstrated that incorporating supervisory signals into SSL can further augment model performance by guiding the learning process with additional relevant information. However, the application of such supervisory signals in the context of disease progression analysis remains largely unexplored. This gap is particularly pronounced given the inherent challenges of incorporating both event and time-to-event information into the learning paradigm. Addressing this, we propose a novel framework, Time and Event-aware SSL (TE-SSL), which integrates time-to-event and event and data as supervisory signals to refine the learning process. Our comparative analysis with existing SSL-based methods in the downstream task of survival analysis shows superior performance across standard metrics. The full code can be found here: \href{https://github.com/jacob-thrasher/TE-SSL}{https://github.com/jacob-thrasher/TE-SSL} \let\thefootnote\relax\footnote{*Data used in preparation of this article were obtained from the Alzheimer’s Disease Neuroimaging Initiative
(ADNI) database (adni.loni.usc.edu). As such, the investigators within the ADNI contributed to the design
and implementation of ADNI and/or provided data but did not participate in analysis or writing of this report.
A complete listing of ADNI investigators can be found 
\href{http://adni.loni.usc.edu/wp-content/uploads/how_to_apply/ADNI_Acknowledgement_List.pdf}{here}}

\end{abstract}

\section{Introduction}

Advancement in deep-learning-based medical image analysis have shown remarkable promise in revolutionizing the study of Alzeimer's Dementia (AD) through medical imaging \cite{wen2020convolutional,arya2023systematic,warren2023functional,chang2023mri}, potentially aiding in the management and treatments of the estimated 6.7 million people aged 65+ who live with AD in the United States \cite{AD2023Stats}. 
For instance, Chang
\textit{et al.} (2023) \cite{chang2023mri} developed MRI-based deep learning framework for differentiating between Alzheimer's disease, temporal lobe epilepsy and healthy controls.  
These approaches have been shown to be successful in identifying subtle patterns in brain images that may be challenging for human detection. 
While significant strides have been made in utilizing computer-assisted tools for detecting Alzheimer's Disease, the focus on progression analysis, such as time-to-event prediction, has been comparatively less prominent. However, the ability to accurately predict the trajectory of Alzheimer's over time is crucial for early diagnosis and can dramatically transform the clinical workflow. Tools capable of forecasting disease progression offer invaluable insights for personalized patient care, enabling timely interventions and better management of the disease's impact on patients' lives.

Representation learning within deep learning has demonstrated remarkable potential in extracting meaningful features from data, significantly enhancing the performance of downstream tasks \cite{bengio2013representation,bachman2019learning,misra2020self}. This approach has been particularly transformative in medical imaging, where it has contributed to breakthroughs in disease detection and diagnosis \cite{krishnan2022self}. Self-supervised learning (SSL) and its variants stand out as key strategies in learning these useful representations without relying on labeled data \cite{misra2020self,chen2020simple,zbontar2021barlow}. When labels are available, incorporating them into SSL frameworks can further refine model performance by effectively clustering data points from the same class closer together while distancing those from different classes \cite{khosla2020supervised}. Despite its success across domains, including medical image analysis \cite{yang2021self,dong2021self,rettenberger2023self}, the application of SSL in analyzing disease progression, such as in Alzheimer's disease, remains under explored. Furthermore, there appears to be a significant gap in the literature, as no existing studies have leveraged the supervisory signals provided by available labels, such as event indicators, to enhance the capabilities of self-supervised learning models specifically for progression analysis in Alzheimer's diseases. Our research aims to bridge this gap by integrating these supervisory signals into our self-supervised learning framework, thereby potentially improving downstream tasks in progression analysis. 

Toward this goal, we initially explored the potential of self-supervised learning (SSL) in the progression analysis of Alzheimer's disease. As anticipated, the initial results showcased improved outcomes that further encouraged us to incorporate supervisory signals to enhance representation learning for AD progression analysis. We integrated the event label to better guide the representation learning specifically tailored for progression analysis. Recognizing the significance of time-to-event information, we developed a novel self-supervised learning framework, Time and Event-aware SSL (TE-SSL), which incorporates both event and time-to-event labels as additional supervisory signals. Through evaluation on the ADNI dataset, TE-SSL demonstrated an improvement in the downstream performance of time-to-event prediction. Overall, our contributions are: 
\begin{enumerate}
    \item The use of supervisory signals in the form of event occurrence for progression analysis, offering a novel approach to understanding disease dynamics.
    \item A novel framework, TE-SSL, that uses time and event labels for SSL training.
    \item Demonstrated improved downstream performance for time-to-event prediction across different metrics, showcasing the practical efficacy of our approach in enhancing the predictive capabilities for Alzheimer's disease progression.
\end{enumerate}

\section{Methods}
\begin{figure}[t]
    \centering
    \includegraphics[width=\textwidth]{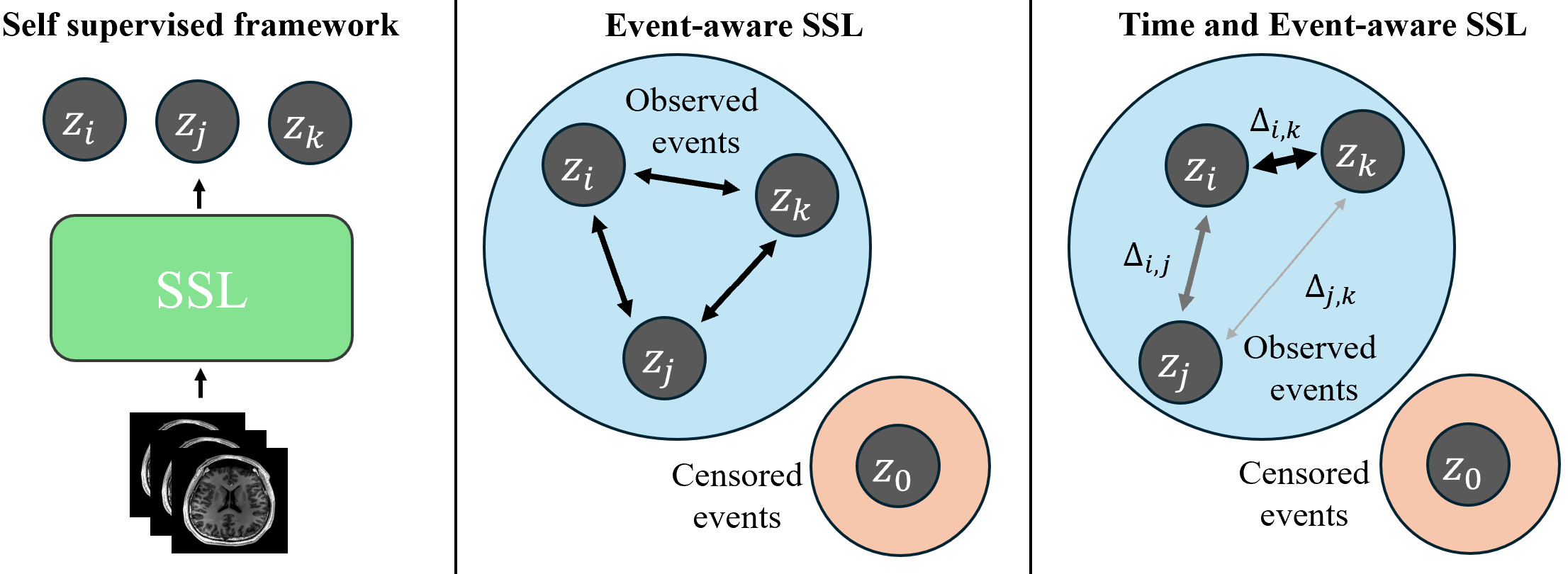}
    \caption{Schematic diagram of the proposed time- and event-aware SSL, where $\Delta_{*,*}$ represents the time difference between two data elements $z_*$}
    \label{fig:schematic}
\end{figure}

We consider a set of labeled training examples $\mathcal{X}$  with the corresponding labels $\mathcal{Y}$. Since, we are interested in modeling disease progression, we consider a setup of survival analysis, where the labels  $\mathcal{Y}$ contains both time and event information. 
Specifically, for each instance $i$, $\mathcal{Y}_i = (T_i, \delta_i)$, where $T_i$ denotes the time to event or censoring and $\delta_i$ is the event indicator, with $\delta_i = 1$ if the event (disease progression) occurred, and $\delta_i = 0$ if the data is censored. Here, censoring refers to instances where the event \textit{has not yet occurred.} Importantly, this does not mean the event will never occur, only that it was not observed during the study. The goal of survival analysis is to model the survival function $S(t) = P(T > t)$, which estimates the probability of an event not occurring by time $t$. 
Building upon this setup, we first learn appropriate representations using our proposed self-supervised learning approach, TE-SSL (section \ref{tessl}), which is a contrastive learning paradigm that utilizes the additional time labels afforded by a survival analysis to strengthen the pull of elements at similar stages of development to improve feature extraction. Fig. \ref{fig:schematic} provides an illustration of our proposed SSL framework. 
We then finetune the network with a task-specific objective function to predict survival outcomes at individual time points (section \ref{downstream}). 

\subsection{TE-SSL: Time and Event-aware Self Supervised
Learning}
\label{tessl}

We provide a background of contrastive self-supervised learning, a learning paradigm designed to use unlabeled data to learn representations by performing some pretext task. 
We then provide details for enhancing the SSL learning paradigm by introducing supervisory signals, before finally laying out our proposed TE-SSL which leverage both event and time-to-event
labels. 


\subsubsection{Self-supervised learning}
Self-supervised learning (SSL) is an unsupervised method for learning representations through pretext tasks, without labeled data. Formally, given training examples $\mathcal{X}$, they are transformed into a modified version $\tilde{\mathcal{X}}$ using a set of transformations $\mathcal{T}$ to create multi-viewed examples. Transformations $\mathcal{T}$ include a set of augmentation functions that do not alter the intrinsic information contained within the data, thus creating a separate view of the original data. SSL frameworks are trained with these multi-viewed batches. For each index $i \in I \equiv {1, 2, ..., 2N}$ in such a multi-viewed batch, let $j$ represent the corresponding index of the transformed pair to sample $i$. In such a setup, a loss function for the self-supervised learning objective can be represented as:
\begin{equation}
\label{eq:ssl}
    \mathcal{L}_{\text{SSL}} = -\sum_{i \in I} \log{\frac{\exp{(z_i \cdot z_j/\tau)}}{\sum_{a \in A(i)} \exp{(z_i} \cdot z_a / \tau)}}
\end{equation}
where $z_*$ is the corresponding projection of the input $\mathbf{x}$, $\tau$ is a temperature parameter, and $A(i) \equiv I - \{i\}$. 
Since augmented images are semantically identical to one another, this method provides anchor points for the model to "pull" together two augmented views while "pushing" other samples in a batch.

\subsubsection{Supervisory signal for SSL training}
The standard SSL setup described above cannot maximize similarity between samples of the same class (e.g., disease category or event vs. censored) in a batch due to the absence of a supervisory signal. While it may seem that standard SSL could eventually differentiate such attributes over time, recent work has shown that incorporating supervisory signals significantly enhances the standard SSL setup \cite{khosla2020supervised}. 

To incorporate supervisory signals using available label information (in our case, event labels), the SSL objective (in Eqn. \ref{eq:ssl}) can be generalized as follows:
\begin{equation}
\label{eq:e-ssl}
    \mathcal{L}_{\text{E-SSL}} = \sum_{i \in I} \frac{-1}{|P(i)|} \sum_{p \in P(i)} \log{\frac{\exp{(z_i \cdot z_p / \tau)}}{\sum_{a \in A(i)} \exp{(z_i \cdot z_a) / \tau}}}
\end{equation}
where $P(i) \equiv \{p \in A(i) | \tilde{y}_p = \tilde{y}_i\}$ represents the set of indices for all positive samples in the multi-viewed batch that are distinct from $i$, and $|P(i)|$ is the cardinality of this set. If sample $i$ has an event indicator during the study period, this objective classifies all samples with an event indicator as positive cases and those that are censored as negative cases, and vice-versa. Clinically, this would translate to maximizing feature representations of multiple patients' imaging data (within a batch) who eventually convert to AD.




\subsubsection{Time-to-event and Event labels for SSL training}
Progression analysis uniquely benefits from having both event indicators and time-to-event information as labels, providing a comprehensive view of patient outcomes. When considering disease progression, it is reasonable to assume that the features of two patients at similar stages of progression will be more similar than those from a late-stage and early-stage patient pair. We therefore hypothesize that utilizing both types of labels in SSL training enhances the learning process, leading to nuanced models that can accurately predict the event timing and occurrence, thus significantly improving disease progression analysis. 

Towards this, we devise weighing schemes for each sample in the batch relative to the anchor point, based on their time-to-event information. We calculate the time difference between the anchor point and other samples in a batch, using the maximum and minimum differences to assign weights to each pair. These weights determine the strength of feature similarity, enforcing that patients at similar stages in development will have a stronger pull than an early/late stage patient pair.
Our proposed SSL learning objective function incorporating both time-to-event and labels is represented as:
\begin{equation}
\label{eq:te-ssl}
    \mathcal{L}_{\text{TE-SSL}} = \sum_{i \in I} \frac{-1}{|P(i)|} \sum_{p \in P(i)} \log{\frac{\omega_{i, p} \exp{(z_i \cdot z_p / \tau)}}{\sum_{a \in A(i)} \omega_{i, a} \exp{(z_i \cdot z_a) / \tau}}}
\end{equation}
where the weight term $\omega_{*,*}$ for each anchor point is calculated as:
\begin{equation}
\label{eq:weight}
    \omega_{i, j} = \frac{\alpha - \beta}{s - l}\Delta_{i, j} + \frac{\beta - \alpha}{s - l}s + \alpha
\end{equation}
where 
$\Delta_{i, j} = |T_i - T_j|$ is the time difference associated with data points $i$ and $j$. Additionally, we compute $\Lambda = \{\Delta_{i, j} | i, j \in I, i \neq j\}$ and take $s = \min{\Lambda}$ and $l = \max{\Lambda}$ to establish the maximum and minimum time span differences between samples in the batch.
Finally, $\alpha$ and $\beta$ serve as hyperparameters defining the maximum and minimum weight values, respectively. Specifically, the pair with the smallest time difference is assigned the highest weight, $\alpha$, and the pair with the largest time difference receives the lowest weight, $\beta$. 

\subsection{Time-to-event prediction}
\label{downstream}
To leverage the feature space learned with SSL frameworks, we construct a deep learning framework consisting of an encoder network $\mathcal{E}(\cdot)$ and a projection head $\mathcal{P}(\cdot)$. During pretraining (Section \ref{tessl}), multiviewed data (original and it's augmented copy  $\tilde{\mathcal{X}})$ is passed through the encoder module to obtain $r = \mathcal{E}(\tilde{\mathcal{X}}) \in \mathbb{R}^{d_E}$, where $d_E$ is the dimension of $r$. A final representation $z = \mathcal{P}(r) \in  \mathbb{R}^{d_P}$ ($d_P$ is the dimension $z$) is computed and normalized for the pretraining procedure. After pretraining, $\mathcal{P}(\cdot)$ is discarded and replaced with task-specific head, which is then finetuned together with the encoder network on the time-to-event objective.

For our task head, we adopt the DeepHit framework \cite{lee2018deephit}, a deep learning approach to survival analysis. With DeepHit framework, instead of predicting a single hazard coefficient for a given input, we output a distribution of hazards at discrete time points. This allows the model to learn the first hitting times (predicted time until the occurrence of the first event of interest for each subject) directly without making assumptions about the underlying form of the data. 
In specific, the 
model learns to minimize the loss function $\mathcal{L}_{total} = \mathcal{L}_1 + \mathcal{L}_2$, where $\mathcal{L}_1$ is the log-likelihood of the distribution of the hitting time, defined as

\begin{equation}
    \mathcal{L}_1 = -\sum^{N}_{i=1}[\mathbbm{1}(\delta_i = 1) * \log{h_i^{T_i}} + \mathbbm{1}(\delta_i \neq 1) * \log{(1 - \hat{F}(T_i|x_i)}]
\end{equation}
where, $\mathbbm{1}$ is an indicator function evaluating to 1 iff $\delta_i = 1$ (event occurred), $h_i^{T_i}$ corresponds to the predicted hazard for input $X_i$ at time $T_i$, and $\hat{F}(T_i|x_i)$ is the estimated cumulative incidence function (CIF) which approximates the probability that the event will occur on or before time $T_i$. $\mathcal{L}_2$ incorporates a combination of cause-specific
ranking loss functions and is defined as:
\begin{equation}
    \mathcal{L}_2 = \gamma \sum_{i \neq j} A_{i, j} \cdot \exp (\hat{F}(T_i | x_i), \hat{F}(T_i | x_j))
\end{equation}
where $\gamma$ is a hyperparameter which indicates the intensity of the ranking loss and $A_{i, j} = \mathbbm{1}(T_i < T_j)$ represents an indicator function which evaluates to $1$ if a pair $(i, j)$ experience an event at different times.


\section{Experiments and Results}

\subsubsection{ADNI dataset:} Our data consists of a cohort of 493 unique patients in the Alzheimer's Disease Neuroimaging Initiative (ADNI) \cite{Petersen} dataset. Each subject has one or more visits containing a 3D T1-weighted MR Image, yielding a total of 2007 data points. Patients are diagnosed as being cognitively normal (CN), having mild cognitive impairment (MCI), or Alzheimer's dementia (AD) at every visit. We define \textit{converters} as subjects whom were CN or MCI during their initial visit, but developed AD within the duration of the study. Additionally, each visit contains the number of months since the baseline observation, which acts as the time-to-event signal.
The data were preprocessed via the pipeline laid out by \cite{liu2020} and divided based on the unique participants to avoid data leakage. For patients with multiple visits, we treat each visit as a unique data point.

\subsubsection{Implementation details:}
We utilize a 3D CNN adapted from \cite{liu2020} as our backbone MRI encoder for both pretraining and finetuning tasks. The encoder takes in $X \in \mathbb{R}^{N \times 96 \times 96 \times 96}$ and outputs a representation $\mathcal{E}(X) = r \in \mathbb{R}^{N \times 1024}$, where $N$ is the batch size.

\textit{Pretrain phase:} Contrastive based SSL techniques require large batch sizes to train properly. Due to hardware constraints, we selected $N=16$ and accumulated gradients for 8 iterations before backpropagation to simulate a batch size of 128. The encoded representation $r$ is then passed through the projection head to achieve $z = \mathcal{P}(r) \in \mathbb{R}^{N \times 128}$. We assigned a temperature of $\tau = 0.07$ for all contrastive loss functions and optimized the model using LARS with a learning rate of $.3 \times N/256 = .15$ \cite{Chen2020} and a momentum of 0.9.

 \textit{Finetune phase:} We first discard the projection head of the pretrained model and replace it with a task-specific $\mathcal{P}_t(\cdot)$ to predict the probability mass function (PMF) associated with the input. We then apply the Adam optimizer with a learning rate of $1e-4$ during training. 

 \textit{Evaluation:} We compute the Time-dependent Concordance Index (C-td) \cite{antolini2005time}, which measures the extent to which the ordering of actual survival times of pairs agrees with the ordering of their predicted risk. Additionally, we evaluate the Integrated Brier Scores (IBS) \cite{graf1999assessment}, which measures an overall assessment of the model's performance across all available times considered in the study. These metrics are calculated using the Pycox library. 

\begin{table}[!t]
\caption{Comparison between different time-to-event prediction approach for C-td $\uparrow$ and IBS $\downarrow$ scores. 
\textbf{Bold} and \underline{underline} denote 1st and 2nd best values, respectively.
}\label{tab_overall}
\begin{center}
\begin{tabular}{|c|c|c|}
\hline
Methods                                 & C-td $\uparrow$   & IBS $\downarrow$      \\ \hline
No pretraining   &  0.7329            & 0.2099                 \\
SSL                                     & 0.7511            & \underline{0.1985}                 \\
E-SSL                                   & \underline{0.7720}            & 0.1997                 \\
TE-SSL & \textbf{0.7873} & \textbf{0.1889} \\
\hline                                               
\end{tabular}
\end{center}
\end{table}

\begin{figure}[t]
    \centering
\includegraphics[width=\textwidth]{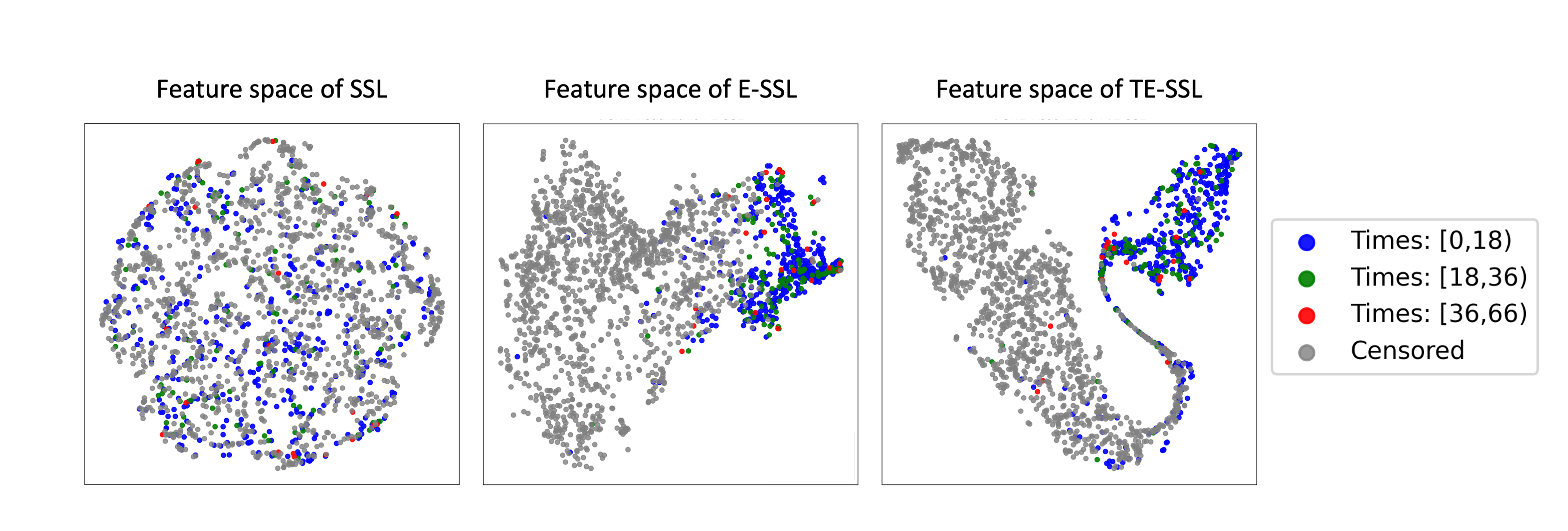}
    \caption{t-SNE analysis of representations across different SSL frameworks. Individual points, if not censored, are labeled with different time-to-event groups.
    }
    \label{fig:tsne}
\end{figure}

\subsection{Results}
Our primary results in Table \ref{tab_overall} showcase a comparison between our E-SSL and TE-SSL frameworks and baseline models: \textit{No Pretraining} and \textit{SSL}. For fairness, all models, including E-SSL and TE-SSL, were finetuned using the same model as \textit{No Pretraining}. We trained each model with three random seeds and reported their average results. Our frameworks outperform others in both C-td (higher is better) and IBS (lower is better) metrics. Notably, E-SSL introduces the novel use of event labels in SSL training for progression analysis, while TE-SSL's innovative incorporation of both time-to-event and event information leads to the best performance, highlighting its efficacy in progression analysis.

In Fig. \ref{fig:tsne}, we present t-SNE plots \cite{van2008visualizing} to examine the feature spaces learned by the SSL frameworks considered in this study. The left panel displays the feature space resulting from standard SSL training, which lacks the clear separation between censored patient data and patient data with events observed in the middle and right panels, where supervisory signals enhance SSL training. Among the two SSL frameworks with supervisory signals, the incorporation of both time-to-event and event information appears to achieve superior separability.

\subsection{Ablation analysis}
We conducted the ablation analysis to better understand the roles of $\alpha$ and $\beta$ in our proposed TE-SSL (Eqn. \ref{eq:weight}), which serve to define the intensity of the weight values for a pair of inputs $(i, j)$. 
The difference between $\alpha$ and  $\beta$ dictates how strongly to differentiate distant pairs. 
For instance, in a batch of $N$ samples if $(i, j)$ represents the pair with the largest time difference, setting $\beta = 0$, effectively considers $j$ as a negative sample relative to the anchor $i$. Therefore, we explored sensible configurations of $\alpha$ and $\beta$, with $1 \leq \alpha \leq 1.5$ and $0.5 \leq \beta \leq 1$, noting that an $\alpha - \beta < 0.5$ would inappropriately diminish the negative impact of distant pairs. The results, presented in Table \ref{tab_alpha_beta_variation}, demonstrate the method's relative stability within these selected ranges. It is also noteworthy that TE-SSL outperforms both standard SSL and the baseline no-pretraining time-to-event prediction model in five out of six experiments, highlighting its efficacy.

\begin{table}[!t]
\caption{Model performance on different values of $\alpha$ and $\beta$ for TE-SSL pretraining. 
}\label{tab_alpha_beta_variation}
\begin{center}
\begin{tabular}{|c|c|c|c|}
\hline
$\alpha$   & $\beta$              & C-td $\uparrow$     & IBS  $\downarrow$\\ \hline
1                           & 0.5                       & 0.7636                        & 0.1954                     \\
1                           & 0.7                       & 0.7581                        & 0.1954                     \\
1                           & 0.9                       & \textbf{0.7883}               & \textbf{0.1889}            \\
1.1                         & 1                         & \underline{0.7714}            & \underline{0.1981}               \\
1.3                         & 1                         & 0.761                         & 0.1931                     \\
1.5                         & 1                         & 0.733                         & 0.2027      
\\ \hline 
\end{tabular}
\end{center}
\end{table}


\section{Conclusion}
We introduce the Time and Event-aware SSL framework, which integrates both event and time-to-event information to guide the learning process of feature representations. As demonstrated, our approach surpasses existing self-supervised learning methods, including those supervised versions that incorporate only the event label. This underscores the critical importance of utilizing both event and time-to-event information in the progression analysis of Alzheimer's disease. Our evaluation using the ADNI dataset showcases the practical applicability and effectiveness of our proposed method, significantly contributing to the advancement of AD progression study. 

\subsubsection{Acknowledgments.} 
This research was supported by 
West Virginia Higher Education Policy Commission's Research Challenge Grant Program 2023 and 
DARPA/FIU AI-CRAFT grant. 
Data collection and sharing for the Alzheimer's Disease Neuroimaging Initiative (ADNI) is funded by the National
Institute on Aging (National Institutes of Health Grant U19 AG024904). The grantee organization is the Northern
California Institute for Research and Education. In the past, ADNI has also received funding from the National
Institute of Biomedical Imaging and Bioengineering, the Canadian Institutes of Health Research, and private
sector contributions through the Foundation for the National Institutes of Health (FNIH) including generous
contributions from the following: AbbVie, Alzheimer’s Association; Alzheimer’s Drug Discovery Foundation;
Araclon Biotech; BioClinica, Inc.; Biogen; Bristol-Myers Squibb Company; CereSpir, Inc.; Cogstate; Eisai Inc.;
Elan Pharmaceuticals, Inc.; Eli Lilly and Company; EuroImmun; F. Hoffmann-La Roche Ltd and its affiliated
company Genentech, Inc.; Fujirebio; GE Healthcare; IXICO Ltd.; Janssen Alzheimer Immunotherapy Research \&
Development, LLC.; Johnson \& Johnson Pharmaceutical Research \& Development LLC.; Lumosity; Lundbeck;
Merck \& Co., Inc.; Meso Scale Diagnostics, LLC.; NeuroRx Research; Neurotrack Technologies; Novartis
Pharmaceuticals Corporation; Pfizer Inc.; Piramal Imaging; Servier; Takeda Pharmaceutical Company; and
Transition Therapeutics. 

\subsubsection{Disclosure of interests.} The authors have no competing interests to report.

\bibliography{export}
\bibliographystyle{plain}
\end{document}